\def\year{2022}\relax
\documentclass[letterpaper]{article} 
\usepackage{aaai22}  
\usepackage{times}  
\usepackage{helvet}  
\usepackage{courier}  
\usepackage[hyphens]{url}  
\usepackage{graphicx} 
\usepackage{natbib}  
\usepackage{caption} 
\DeclareCaptionStyle{ruled}{labelfont=normalfont,labelsep=colon,strut=off} 
\usepackage{algorithm}

\usepackage[switch]{lineno}
\usepackage{amssymb}
\usepackage{latexsym}
\usepackage{mathtools}
\usepackage{algorithm}
\usepackage[noend]{algpseudocode}
\usepackage{setspace}

\usepackage{graphicx}
\usepackage{amsmath}
\usepackage{amsthm}
\usepackage{amssymb}
\usepackage{booktabs}
\usepackage{algorithm}
\usepackage{array}
\usepackage{subfigure}
\usepackage{multirow}
\usepackage{threeparttable}
\usepackage{tabularx}
\usepackage{booktabs}
\usepackage{colortbl}
\usepackage{xcolor}
\usepackage{stfloats}
\usepackage[noend]{algpseudocode}
\usepackage{mathtools}
\usepackage{makecell}
\usepackage{setspace}
\usepackage{CJKutf8}

\usepackage{newfloat}
\usepackage{listings}
\floatstyle{ruled}
\newfloat{listing}{tb}{lst}{}
\floatname{listing}{Listing}
\title{MarkBERT: Marking Word Boundaries Improves Chinese BERT}
\author{Linyang Li$^2$\thanks{\ \ \ Work done during internship at Tencent AI Lab.}\ ,Yong Dai$^{1}$,  Duyu Tang$^1$\thanks{~~Corresponding author.}\ \ , \textbf{Xipeng Qiu$^2$, Zenglin Xu$^3$, Shuming Shi$^1$} \\
}
\affiliations{
	$^1$ Tencent AI Lab, China,$^2$ Fudan University,$^3$ PengCheng Laboratory \\
\{yongdai,duyutang\}@tencent.com,  \\
 \{linyangli19, xpqiu\}@fudan.edu.cn
}

\usepackage{bibentry}

\begin{document}
\begin{CJK*}{UTF8}{gbsn}
\maketitle

\begin{abstract}
We present a Chinese BERT model dubbed MarkBERT that uses word information in this work.
Existing word-based BERT models regard words as basic units, however,
due to the vocabulary limit of BERT, they only cover high-frequency words and fall back to character level when encountering out-of-vocabulary (OOV) words.
Different from existing works, MarkBERT keeps the vocabulary being Chinese characters and inserts boundary markers between contiguous words. 
Such design enables the model to handle any words in the same way, no matter they are OOV words or not.
Besides, our model has two additional benefits:
first, it is convenient to add word-level learning objectives over markers, which is complementary to traditional character and sentence-level pretraining tasks;
second, it can easily incorporate richer semantics such as POS tags of words by replacing generic markers with POS tag-specific markers.
With the simple markers insertion, MarkBERT can improve the performances of various downstream tasks including language understanding and sequence labeling.
\footnote{All the codes and models will be made publicly available at \url{https://github.com/daiyongya/markbert}}

\end{abstract}

\section{Introduction}

Chinese words can be composed of multiple Chinese characters. 
For instance, the word 地球 \ (earth) is made up of two characters 地 \ (ground) and 球 \  (ball).
However, there are no delimiters (i.e., space) between words in written Chinese sentences.
Traditionally, word segmentation is an important first step for Chinese natural language processing tasks \cite{chang2008optimizing}. 
Instead, with the rise of pretrained models \cite{bert}, Chinese BERT models are dominated by character-based ones \cite{cui2019pre,sun2019ernie,cui-etal-2020-revisiting,sun2021chinesebert,sun2021ernie}, where a sentence is represented as a sequence of characters.
There are several attempts at building Chinese BERT models where word information is considered. 
Existing studies tokenize a word as a basic unit \cite{zhuiyiwobert}, as multiple characters \cite{cui2019pre} or a combination of both \cite{zhang2020ambert,lai2021lattice,Guo2021LICHEEIL}. 
However, due to the limit of the vocabulary size of BERT, these models only learn for a limited number (e.g., 40K) of words with high frequency. 
Rare words below the frequency threshold will be tokenized as separate characters so that the word information is neglected.

In this work, we present a simple framework, MarkBERT, that considers Chinese word information. 
Instead of regarding words as basic units, we use character-level tokenizations 
and inject word information via inserting special markers between contiguous words. 
The occurrence of a marker gives the model a hint that its previous character is the end of a word and the following character is the beginning of another word.
Such a simple model design has the following advantages.
First, it avoids the problem of OOV words since it deals with common words and rare words (even the words never seen in the pretraining data) in the same way.
Second, the introduction of marker allows us to design word-level pretraining tasks (such as replaced word detection illustrated in section \ref{sec:markbert}), which are complementary to traditional character-level pretraining tasks like masked language modeling and sentence-level pretraining tasks like next sentence prediction.

In the pretraining stage, 
we force the markers to understand the contexts around them while serving as separators between words.
We train our model with two pretraining tasks.
The first task is 
masked language modeling and we also mask markers such that word boundary knowledge can be learned since the pre-trained model needs to recognize the word boundaries within the context.
The second task is replaced word detection.
We replace a word with artificially generated words and ask the markers behind the word to predict whether the word is replace.
Such a process will force the markers to serve as discriminators therefore can learn more word-boundary information within the context.
With these two pretraining tasks, we train the MarkBERT model initialized from BERT-Chinese models and obtain considerable improvements.

We conduct extensive experiments on various downstreams tasks including named entity recognition tasks (NER) and natural language understanding tasks.
On the NER task, we demonstrate that MarkBERT can significantly surpass baseline methods on both MSRA and OntoNotes datasets \cite{huang2015bidirectional,zhang2018chinese}.
Compared with other word-level Chinese BERT models, we conduct experiments and observe that MarkBERT performs better on text classification, keyword recognition, and semantic similarity tasks in the CLUE benchmark datasets. 

We summarize the major contributions of this work as follows.

\begin{itemize}
    \item We present a simple and effective Chinese pretrained model MarkBERT that considers word information without aggravating the problem of OOV words.
    \item We demonstrate that our model achieves considerable performance improvements on Chinese NER and Chinese NLU tasks with a simple yet effective mark insertion strategy.
\end{itemize}

\begin{figure*}[]
\centering
\includegraphics[width=1.0\linewidth]{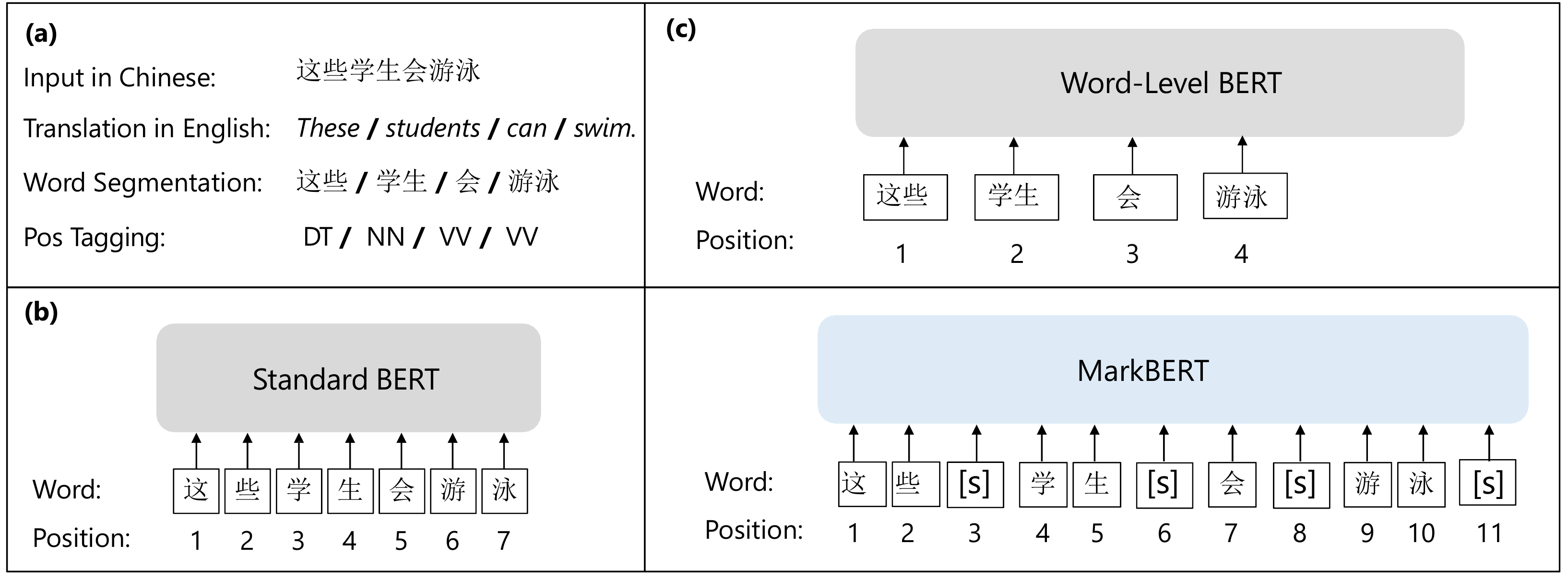}
\centering
\caption{An illustrative example of our model. Box (a) gives the original input written in Chinese, its translation in English,  word segmentation results given by an off-the-shell text analyzer, and the POS tags of words. Box (b) shows a traditional character-level Chinese BERT. Box (c) shows a word-level BERT using word-level vocabulary in the encoding process.
In box (d), we show the structure of MarkBERT which inserts markers \texttt{[S]} between words but the model remains a character-level model.}
\label{fig:illustration}
\end{figure*}

\section{Related Work}
We describe related work on injecting word information to Chinese BERT and the use of marker in natural language understanding tasks. 
\subsection{Chinese BERT}

Pre-trained models exemplified by BERT \cite{bert} and RoBERTa \cite{cui2019pre} have been proved successful in various Chinese NLP tasks \cite{xu2020clue, cui-emnlp2019-cmrc2018}.
Existing Chinese BERT models that incorporate word information can be divided into two categories.
The first category uses word information in the pretraining stage but represents a text as a sequence of characters when the pretrained model is applied to downstream tasks. 
For example, \citet{cui2019pre} use the whole-word-masking strategy that masks word spans and predicts continuously multiple masked positions.
\citet{lai2021lattice} incorporate lexicon information by concatenating the lexicons along with character-level context.
The second category uses word information when the pretrained model is used in downstream tasks.  
For example, \citet{zhuiyiwobert} uses a word-level vocabulary instead of characters.
If a word 地球 \ is included in the vocabulary, its constitutes 地 \ and 球 \ will not be considered as input tokens.
\citet{zhang2020ambert} go one step further by constructing two independent encoders that encode character-level and word-level information separately and concatenate them at the top layers of two encoders.
Similarly, \citet{Guo2021LICHEEIL} encode both character-level and word-level information. They move the information aggregation stage to the embedding level.


\subsection{Marker Insertion in NLU}

The idea of inserting markers is explored in entity-related natural language understanding tasks, especially in relation classification. 
Given a subject entity and an object entity as the input, existing work inject untyped markers \cite{sun2019ernie,soares2019matching} or entity-specific markers \cite{zhong2020frustratingly} around the entities, and
make better predictions of the relations of the entities.

\section{MarkBERT Pre-training}

In this section, we first introduce the background of character level Chinese pre-trained models;
then we introduce the structure of our MarkBERT model.
After describing the structure of MarkBERT, we introduce the training process of the MarkBERT.
Finally, we provide details of the entire training process.

\label{sec:markbert}

\subsection{Character Level Chinese BERT}

In language model pre-training, BERT \cite{bert} first introduced the masked language modeling strategy to learn the context information by replacing tokens with masks and assign the model to predict the masked tokens based on the contexts around them using the self-attention transformers structure \cite{vaswani2017attention}.
In Chinese language model pre-training, the encoding unit is different from the widely used BPE encoding in English: Chinese pre-trained models are usually character-level and 
word level information is typically neglected.

\begin{figure*}[]
\centering
\includegraphics[width=1.0\linewidth]{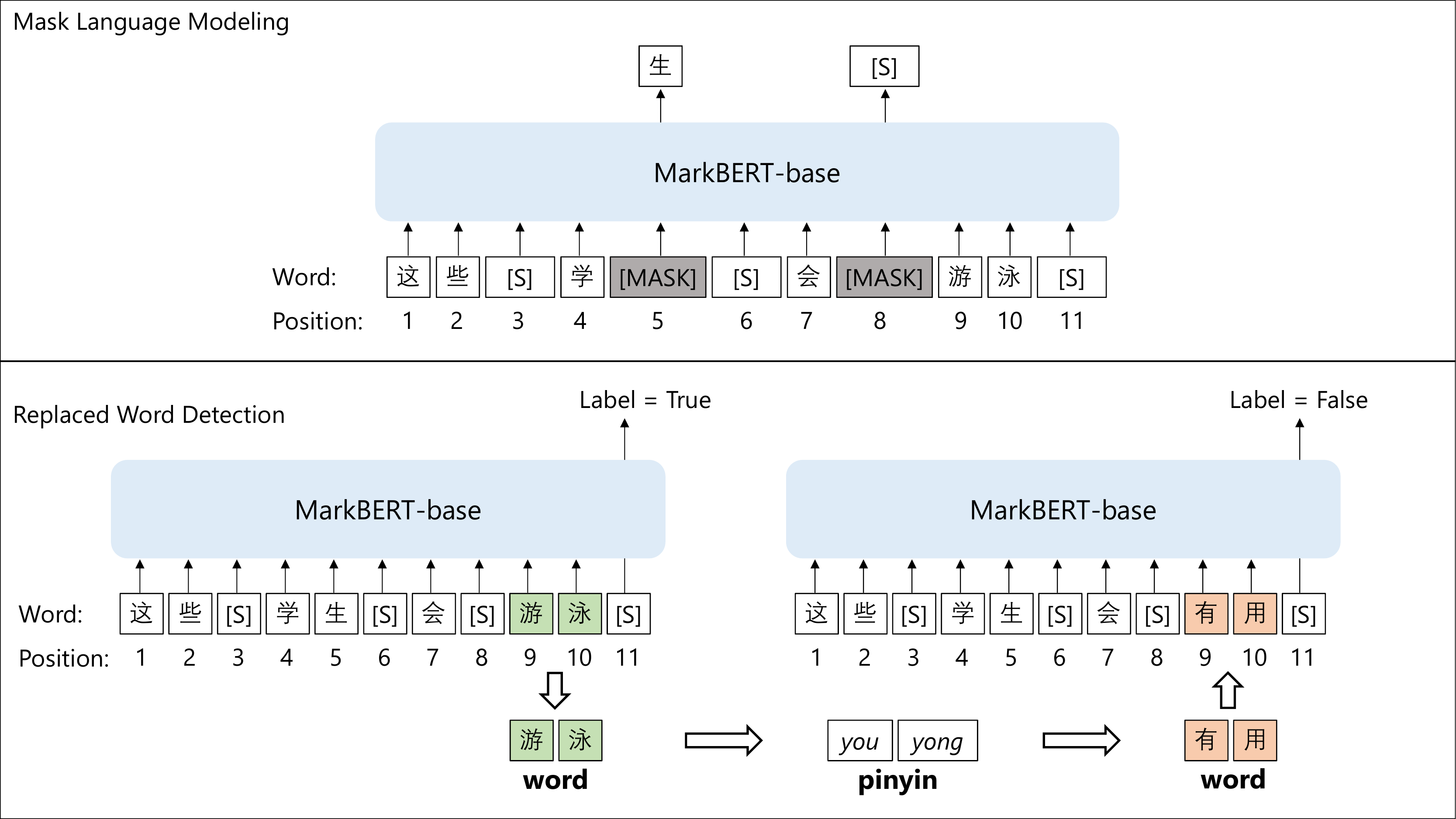}
\centering
\caption{Illustration of the predicting tasks of Masked Language Modeling and Replaced Word Detection. Here, \texttt{[S]} is the inserted markers.}
\label{fig:main}
\end{figure*}

\subsection{MarkBERT Model}

To make better use of word-level information in Chinese pre-training, we introduce a simple framework called MarkBERT. 
We insert markers between word spans to give explicit boundary information for the model pre-training.

As seen in Figure \ref{fig:illustration}, we first use a segmentation tool to obtain word segmentations, then we insert special markers between word spans as separators between characters.
These markers are treated as normal characters so they take positions in the transformers structure.
Plus, they can also be masked for the mask language modeling task to predict, 
therefore the encoding process needs to be aware of predicting word boundaries rather than simply filling in masks from the context.
The mask prediction task becomes more challenging since predicting the masks correctly requires a better understanding of the word boundaries. 
In this way, the model is still character-level encoded while it is aware of word boundaries since word-level information is given explicitly.

\subsection{Replaced Word Detection}

Inserting special markers allows the pre-trained model to recognize word boundaries while maintaining a character-level model.
Further, these special markers can be used to construct a word-level pre-training task which can be complementary to the character-level masked language modeling task.

We construct a replaced word detection task as an auxiliary task to the masked language modeling task.
We construct a bipolar classification task that detects whether the word span is replaced by a confusion word.
Specifically, given a word span, we take the representations of the marker after it and make binary prediction.

When a word span is replaced by a confusion word, as seen in Figure \ref{fig:main}, the marker is supposed to make a "replaced" prediction labeled as "False".
When the word spans are not changed, the marker will make an "unchanged" prediction labeled as "True".
Therefore, suppose the representation of the $i^{th}$ marker is $x^i$ with label $y^{true}$ and $y^{false}$, the replaced word detection loss is:
\begin{align}
    \mathcal{L} = - \sum_i [ y^{} \cdot log (x^i) ]
\end{align}
We add this loss term to the masked language modeling loss as a multi task training process.

The construction of the confusions could be various.
We adopt two simple strategies: (1) we use synonyms as confusions; (2) we use words that are similar in phonetics (pinyin) in Chinese.
To obtain the synonyms, we use an external word embedding provided by \citet{zhang2018chinese}.
We calculate the cosine similarity between words and use the most similar ones as the synonyms confusions.
To obtain the phonetic-based confusions, as seen in Figure \ref{fig:main}, we use an external tool to get the phonetics of the word and select a word that share the same phonetics as its confusions.

In this way, the markers can be more sensitive to the word span in the context since these markers are assigned to discriminate the representation type of the word spans before them.
This process is similar to an ELECTRA \cite{clark2020electra} framework.
MarkBERT uses the inserted markers to run the discrimination process inside the encoder and use external confusions instead of using another generator to build texts for the discriminator.

\subsection{Pre-Training}

The pre-training process is a multi task framework consisting of mask language modeling task and replaced word detection task.

In the masked language modeling task, we employ both the masked language modeling strategy and the whole-word-masking strategy. 
In the replaced word detection task, as seen in Figure \ref{fig:main}, when the word span is replaced by confusion words, the model is supposed to correct the confusions. 
This correction process is similar to MacBERT \cite{cui-etal-2020-revisiting}.
For the confusion generation, we use synonyms and pinyin-based confusions.
The synonyms are obtained by a synonym dictionary based on calculating the cosine similarity between the Chinese word-embeddings provided by \citet{zhang2018chinese}.

In our MarkBERT pre-training, the mask ratio is still 15\% of the total characters. 
For 30\% of the time, we do not insert any markers so that the model can also be used in a no-marker setting which is the vanilla BERT-style model.
For 50\% of the time we run a whole-word-mask prediction and for the rest we run a traditional masked language model prediction.
In the marker insertion, for 30\% of the time, we replace the word span with a phonetic(pinyin)-based confusion or a synonym-based confusion word and the marker will predict a phonetic(pinyin)-confusion marker or a synonym-confusion marker; for the rest of the time, the marker will predict a normal-word marker.

Therefore, we only calculate 15 \% percent of loss on these normal markers to avoid imbalance labels of the marker learning process.
During fine-tuning on downstream tasks, we use the markers in the input texts.
Also, we can save the markers and downgrade the model to a vanilla BERT-style model for easier usage.

\subsection{Implementation Details in Pre-training}

\subsubsection{Pre-training Dataset Usage}

We use a collection of raw Chinese texts containing Chinese wikipedia, Chinese novels, news. The entire data size is around 80B characters.
We use a simple word segmentation tool Texsmart \cite{texsmart2020} to tokenize the raw data and obtain pos-tags.
We use the same data preprocess framework used in BERT \cite{bert} which constructs documents containing multiple sentences with the length of the maximum token limit and randomly pick another document to train the next sentence prediction task.

\subsubsection{Pre-Training Settings}

We initialize our model from the Roberta whole-word-mask model checkpoint provided by \citet{cui2019pre}.
Therefore, we use the same character-level vocabulary in training our boundary-aware model.
We use both whole-word-mask and normal character mask strategies in the language model training since we aim to learn inner connections between characters in the given word which cannot be achieved by whole-word-masking alone.

We train the model with a maximum sequence length of 512 for the entire training time.
With the markers inserted, the actual maximum sequence length is smaller but we maintain the length as 512 to keep coordinated with previous pre-trained models.
We use the ADAM optimizer \cite{kingma2014adam} used in BERT with a batch size 8,192 on 64x Tesla V100 GPUs.
We set the learning rate to 1e-4 with a linear warmup scheduler. 
We run the warmup process for 10k steps and train 100k steps in total.

\section{Experiments}

\subsection{NER Task}

In the NER task, we use the MSRA \cite{levow-2006-third} and Ontonotes \cite{weischedel2013ontonotes} datasets with the same data-split as in \citet{ma2019simplify} and \citet{li2020flat}.

We establish several strong baselines to explore the effectiveness of our MarkBERT.
In language understanding tasks, we compare with the RoBERTa-wwm-ext \cite{cui2019pre} baseline, which is a whole-word-mask trained Chinese pre-trained models.
We also further pre-train the RoBERTa model denoted as RoBERTa (ours) and the WoBERT model denoted as WoBERT (ours) based on our collected data which is the same data used in pre-training MarkBERT to make fair comparisons with our model.
In the NER task, we compare with FLAT-BERT \cite{li2020flat} and Soft-Lexicon \cite{ma2019simplify} which are state-of-the-art models on the NER task which incorporate lexicons in the transformers/LSTM structure.

\subsection{Language Understanding Task}

We also conduct experiments on language understanding tasks.
We use various types of tasks from the CLUE benchmark \cite{xu2020clue}.
We use classification tasks such as TNEWS, IFLYTEK; semantic similarity task (AFQMC); coreference resolution task(WSC); keyword recognition (CSL); natural language inference task (OCNLI).

Besides the BERT-style baselines used in the NER task, we also use the word-level information enhanced models as baselines to make comparisons in the language understanding tasks.
We use:

- WoBERT \cite{zhuiyiwobert}: a word-level Chinese pre-trained model initialized from the BERT BASE pre-trained weights. It has a 60k expanded vocabulary containing commonly used Chinese words.

- AMBERT \cite{zhang2020ambert}: a multi-granularity Chinese pre-trained model with two separated encoders for words and characters. The encoding representation is the character-level representation concatenated by the word-level representation;

- LICHEE \cite{Guo2021LICHEEIL}: a multi-granularity Chinese pre-trained model that incorporates word and character representations at the embedding level. 

- Lattice-BERT \cite{lai2021lattice}: the state-of-the-art multi-granularity model that uses lexicons as word-level knowledge concatenated to the original input context. 

\subsection{Downstream Task Implementations}

We use the FastNLP toolkit \footnote{https://github.com/fastnlp/fastNLP} to implement the NER experiment;
We use the Huggingface Transformers \cite{wolf-etal-2020-transformers} to implement all experiments.

For the NER task, we follow the implementation details given in the Transformers toolkit. \footnote{https://github.com/huggingface/transformers}
For the language understanding tasks, we follow the implementation details used in the CLUE benchmark official website and the fine-tuning hyper-parameters used in Lattice-BERT \cite{lai2021lattice}.

In the NER task, we use the marker-inserted inputs in the MarkBERT since we intend to incorporate the word boundary information in recognizing entities.
We use the model with the best development performance to obtain the test set result.
We make a thorough discussion on this topic in the later section.
In the NER evaluation process, we label the inserted marker with the same label as its former token and follow the standard BMESO evaluation process used in \citet{ma2019simplify,li2020flat}.

In the NLU tasks, we use the CLUE benchmark datasets to test our model. 
For the TNEWS task, we run the raw classification results without using the keywords augmentation which is no longer a natural context.
For the IFLYTEK task, we split the context and use the average of the split texts prediction since the average sequence exceeds the max sequence length.
We leave the experiment results '-' if they are not listed in the official website. \footnote{https://github.com/CLUEbenchmark/CLUE}

\begin{table*}[]\setlength{\tabcolsep}{5pt}
    \small
    \centering   
    \begin{tabular}{lcccccccccccc}
        \toprule

      \textbf{}  & \multicolumn{3}{c}{\bfseries MSRA(Test)} & \multicolumn{3}{c}{\bfseries OntoNotes(Dev)} & \multicolumn{3}{c}{\bfseries OntoNotes(Test)} \\ 
      & Acc. & Recall & F1 & Acc. & Recall & F1 & Acc. & Recall & F1 \\

        \midrule
        \midrule
 
        BERT \cite{bert} &   94.9 & 94.1 & 94.5 & 74.8 & 81.8 & 78.2   & 78.0 & 75.7 & 80.3\\  
        RoBERTa \cite{cui2019pre} & 95.3 & 94.9 & 95.1 & 76.8 & 80.7 & 78.7 & 77.6 & 83.5 & 80.5\\
        FLAT-BERT \cite{li2020flat} & - & - & 96.1 & - & - & - & - & -  & 81.8\\
        Soft-Lexicon \cite{ma2019simplify} & 95.8 & 95.1 & 95.4 &- &- &- & 83.4 & 82.2 & 82.8\\
        \midrule
        RoBERTa (ours) & 95.7 & 94.8 & 95.2 & 80.3 & 76.4 & 78.3 & 78.8 & 83.4 & 81.1 \\
        MarkBERT (ours) & \textbf{96.1} & \textbf{96.0} & \textbf{96.1} & \textbf{81.2} & \textbf{81.4} & \textbf{81.3} & \text{81.7} & \textbf{83.7} & \text{82.7} \\
        
        \bottomrule

    \end{tabular}
    \caption{NER results on the MSRA and OntoNotes dataset.}
    \label{tab:seq-lab}
\end{table*}

\subsection{Results on NER Task}

In Table \ref{tab:seq-lab}, our proposed boundary-aware MarkBERT outperforms all baseline models including pre-trained models and lexicon-enhanced models.

Compared with the baseline methods, our proposed MarkBERT with markers inserted between words can lift performances by a large margin.
We can observe that compared with the baseline method RoBERTa(ours) which uses word-level information by pretraining with the whole-word mask strategy, MarkBERT can significantly improve the performances in all datasets.
When we insert markers using the same tokenization process used in pre-training MarkBERT in fine-tuning the MarkBERT in the NER task, 
we obtain a considerable performance improvement,
indicating that the inserted markers catch some important fine-grained information that helps improve entity understanding.
Further, when compared with previous state-of-the-art methods such as Soft-Lexicon \cite{ma2019simplify} and FLAT \cite{li2020flat} which use a combination of lexicon-enhanced LSTMs/transformers and BERT, our model can also achieve similar performance while we do not incorporate any lexicon information which is essential in Chinese language.

Therefore, we can conclude that MarkBERT can improve the NER task with a simple marker insertion strategy without complex lexicons therefore can be widely used in sequence labeling tasks.

\begin{table*}[]\setlength{\tabcolsep}{7pt}
    \small
    \centering   
    \begin{tabular}{lcccccccccccc}
        \toprule

      \textbf{}  & \multicolumn{6}{c}{\bfseries Datasets}\\ 
      & TNEWS & IFLYTEK & AFQMC & OCNLI & WSC & CSL  \\

        \midrule
        \midrule
        \multicolumn{2}{c}{\bfseries DEVELOPMENT}\\
        \midrule
        
        BERT \cite{bert}  & 56.09 & 60.37 & 74.10 & 74.70 & 79.22 & 81.02 \\
        RoBERTa \cite{cui2019pre} & 57.51 & 60.80 & 73.80 & 75.01 & 82.20 & 81.22  \\
        
        RoBERTa (ours) & 57.95 & 60.85 & 74.58 & 75.32 & 84.02 & 81.85  \\
        WoBERT (ours)  & 57.01 & 61.10 & 72.80 & 75.00 & 82.72 & - \\ 
        
        \midrule
        
        MarkBERT (ours)  & \textbf{58.40} &\text{60.68} & \textbf{74.89} & \textbf{75.88} & \textbf{84.60} & - \\
        
        \midrule
        \midrule
        \multicolumn{2}{c}{\bfseries TEST}\\
        
        \midrule
        BERT \cite{bert}  & 56.58 & 60.29 & 73.70 & - & 62.00 & 80.36  \\
        RoBERTa \cite{cui2019pre} & 56.94 & 60.31 & 74.04 & - & 67.80 & 81.00 \\
        AMBERT \cite{zhang2020ambert} & - & 59.73 & 73.86 & - & 78.27 & 85.70 \\
        LICHEE \cite{Guo2021LICHEEIL} & - & 60.94 &  73.65 & - & 81.03 & 84.51 \\
        \midrule
        BERT \cite{lai2021lattice} &  - & 62.20 & 74.00 & - & 79.30 & 81.60  \\
        Lattice-BERT \cite{lai2021lattice} & - & \textbf{62.90} & 74.80 & - & \textbf{82.40}  & 84.00 \\
        \midrule
        RoBERTa (ours) & 57.42 & 61.00 & 73.63 & 72.67 & 79.86 & 81.83 \\
        MarkBERT (ours) & \textbf{58.05}  &  \text{62.57}& \textbf{74.87}& \textbf{73.06} & \text{81.72} & 
        \textbf{85.73}  \\
        \bottomrule

    \end{tabular}
    \caption{Evaluation results on the language understanding tasks.
}
    \label{tab:main}
\end{table*}

\subsection{Results on Language Understanding}

Table \ref{tab:main} shows that comparing with the RoBERTa model that uses the same pre-training data, MarkBERT is superior in all tasks. This indicates that the learned representations contain more useful information for the downstream task fine-tuning.
The word-level model WoBERT (ours) trained with the same data used in MarkBERT only achieves a slightly higher accuracy in the IFLYTEK dataset which might because the IFLYTEK dataset contains very long texts where word-level model is superior since it can process more contexts while the total sequence lengths of character level and word level model are both 512.

When comparing with previous works that focus on word-level information, MarkBERT achieves higher performances than the multi-grained encoding method AMBERT as well as LICHEE which incorporates word information as an additional embedding.
We can assume that adding word-level information through \textit{horizontal} markers is more effective than \textit{vertically} concatenating word-level information.
When comparing with the LatticeBERT model, our method can still reach a competitive level of performance, 
meanwhile the relative improvements of our model is larger than the improvements of the LatticeBERT model.
Please note that the lexicons used in LatticeBERT training actually contains more segmentation possibilities which can significantly increase the downstream task performance over the word segmentation based methods \cite{zhang2018chinese}.
The basic idea of incorporating lexicons is parallel with the marker insertion framework.
MarkBERT makes use of word-level information in a different perspective.

\begin{table*}[]\setlength{\tabcolsep}{8pt}
    \small
    \centering   
    \begin{tabular}{lcccccccccccc}
        
        \toprule

      \textbf{}  & \multicolumn{5}{c}{\bfseries Datasets}\\ 
      & MSRA & Ontonotes & TNEWS & IFLYTEK & AFQMC \\

        \midrule
        \multicolumn{1}{c}{\bfseries DEVELOPMENT} & F1 & F1 & Acc. & Acc. & Acc.\\
        \midrule

        MarkBERT  & \textbf{96.1} & \textbf{82.7} & \textbf{58.4} & 60.6 & \textbf{74.8} \\
    
        MarkBERT-{rwd}-pho & 95.8 & 81.7 & 58.0 & 60.8 & 74.3 \\
        MarkBERT-{rwd}-syn & 95.8 & 81.7 & 58.0 & 60.9 & 74.5 \\
         MarkBERT-MLM & 95.8 & 81.3 & 58.0 & 60.7 & 74.6 \\ 
        MarkBERT-w/o marker & 95.5 & 79.2 & 58.2 & \textbf{61.0} & 74.5 \\
        RoBERTa (ours) & 95.1 & 78.2 & 57.9 & 60.8 & 74.5 \\

        \bottomrule

    \end{tabular}
    \caption{Ablation Studies on the NER and the language understanding tasks using dev set results.}
    \label{tab:bert-style}
\end{table*}

\subsection{Model Analysis}

In this section, we conduct ablation experiments to explore the effectiveness of each parts in our MarkBERT framework in different tasks.

We test different variants of MarkBERT: 

- MarkBERT-MLM only considers the MLM task without the replaced word detection task; the masked language model will predict masked tokens as well as inserted markers. 

- MarkBERT-rwd is a version that removes phonetics words or synonyms separately in the replaced word detection process.

- MarkBERT-w/o marker is a version that removed markers which is the same as the vanilla BERT model.

\subsubsection{MarkBERT-MLM without RWD}

To explore which parts in MarkBERT is more effective, we conduct an experiment as seen in Table \ref{tab:bert-style}.
We only use the masked language modeling task while inserting markers without using the replaced word detection task.
The model only considers inserted markers and masked language modeling tasks, while the markers will be masked and predicted as well.

As seen, the MarkBERT -MLM model gains significant boost in the NER task, indicating that word boundary information is important in the fine-grained task.

In the CLUE benchmark, the situation becomes different:
in the IFLYTEK task, inserting markers will hurt the model performance which is because the sequence length exceeds the maximum length of the pre-trained model.
Therefore, inserting markers will results in a lost of contexts. 
Generally, inserting markers is important in downstream task fine-tuning.
The explicit word boundary information helps MarkBERT learn better contextualized representations.

\subsubsection{Replaced Word Detection}

We also test the effectiveness of the additional replaced word detection task. 
Specifically, we separate two confusion strategies and use phonetics and synonyms confusions solely. 

As seen in Table \ref{tab:bert-style}, 
when the marker learning only includes phonetic (pinyin) confusions, the performances in the fine-tuning tasks are similar with the MarkBERT -MLM model, indicating that the phonetic confusions have a slight improvement based on the inserted markers.
When the word spans are replaced by synonyms only, the performances are slightly lower than using both phonetic and synonym confusions, indicating that augmentation using various types of confusions is helpful.

\begin{figure}[]
\centering
\subfigure[]{
\begin{minipage}[t]{0.43\linewidth}
\includegraphics[width=1.0\linewidth]{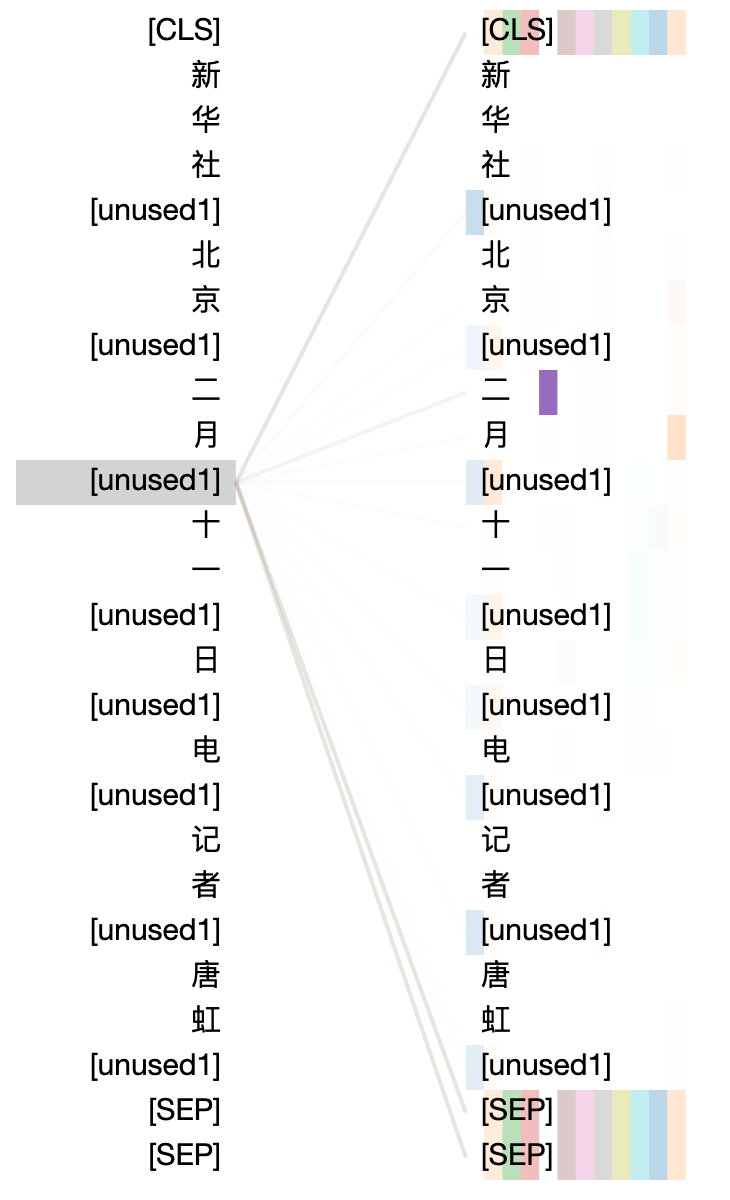}
\end{minipage}%
}%
\subfigure[]{
\begin{minipage}[t]{0.43\linewidth}
\includegraphics[width=1.0\linewidth]{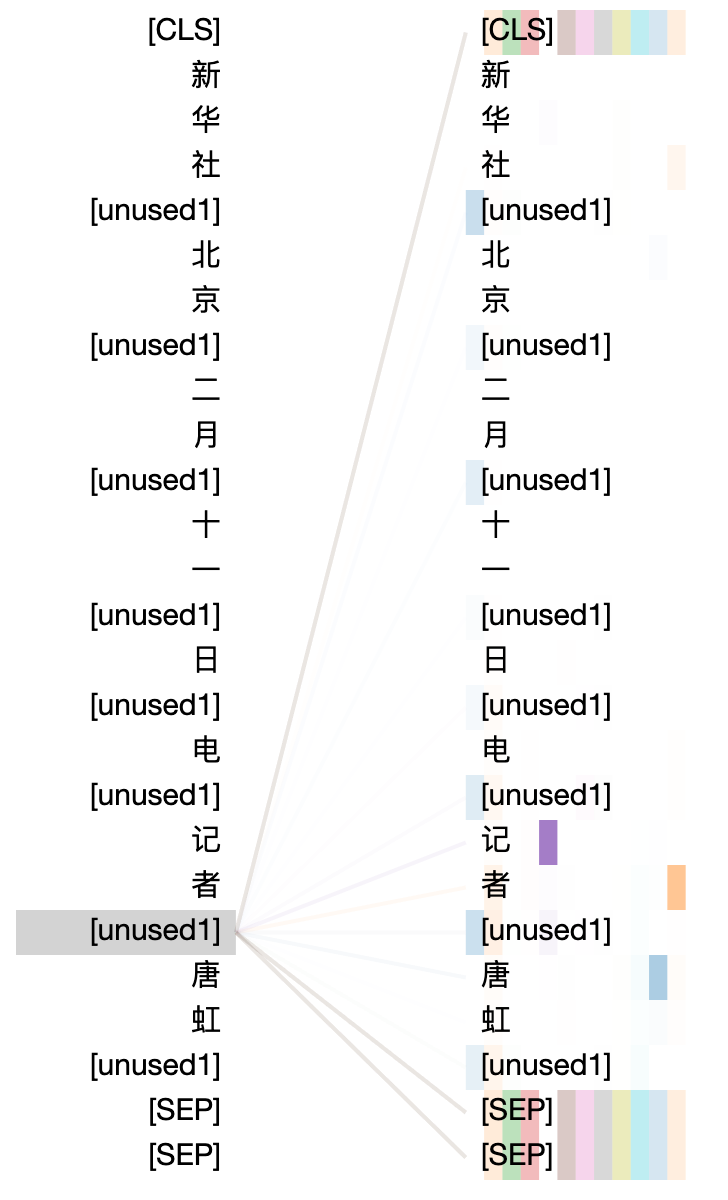}
\end{minipage}%
}%

\subfigure[]{
\begin{minipage}[t]{0.43\linewidth}
\includegraphics[width=1.0\linewidth]{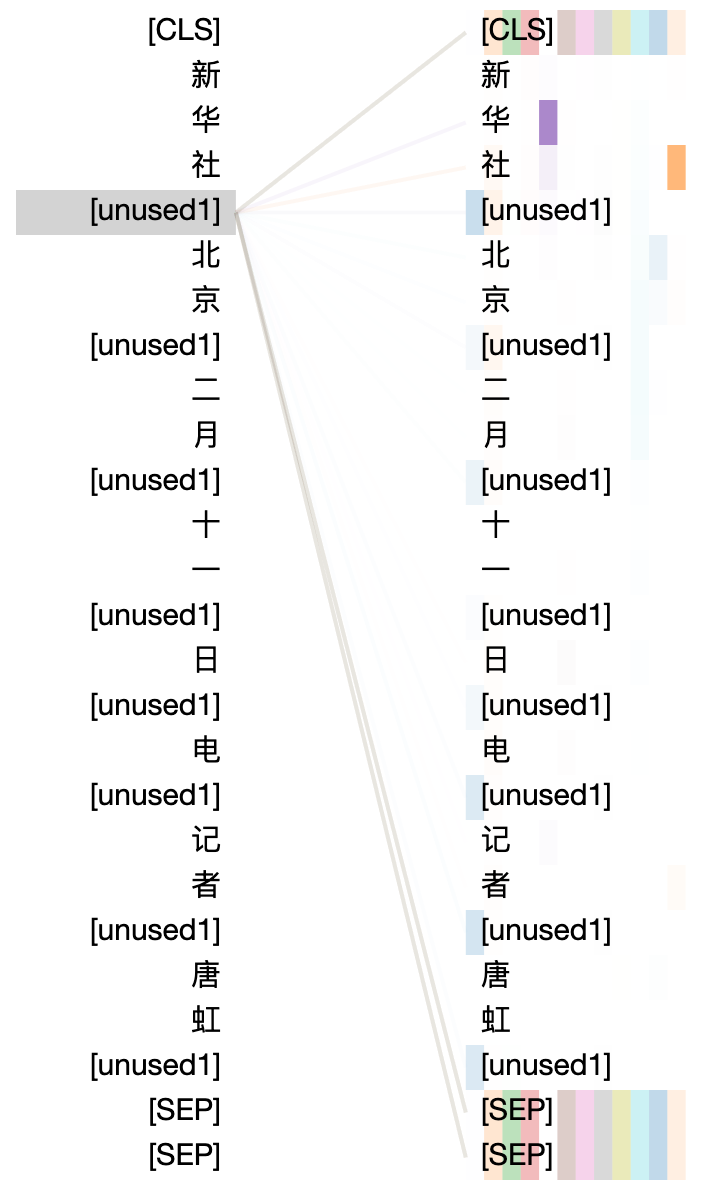}
\end{minipage}%
}%
\subfigure[]{
\begin{minipage}[t]{0.43\linewidth}
\includegraphics[width=0.97\linewidth]{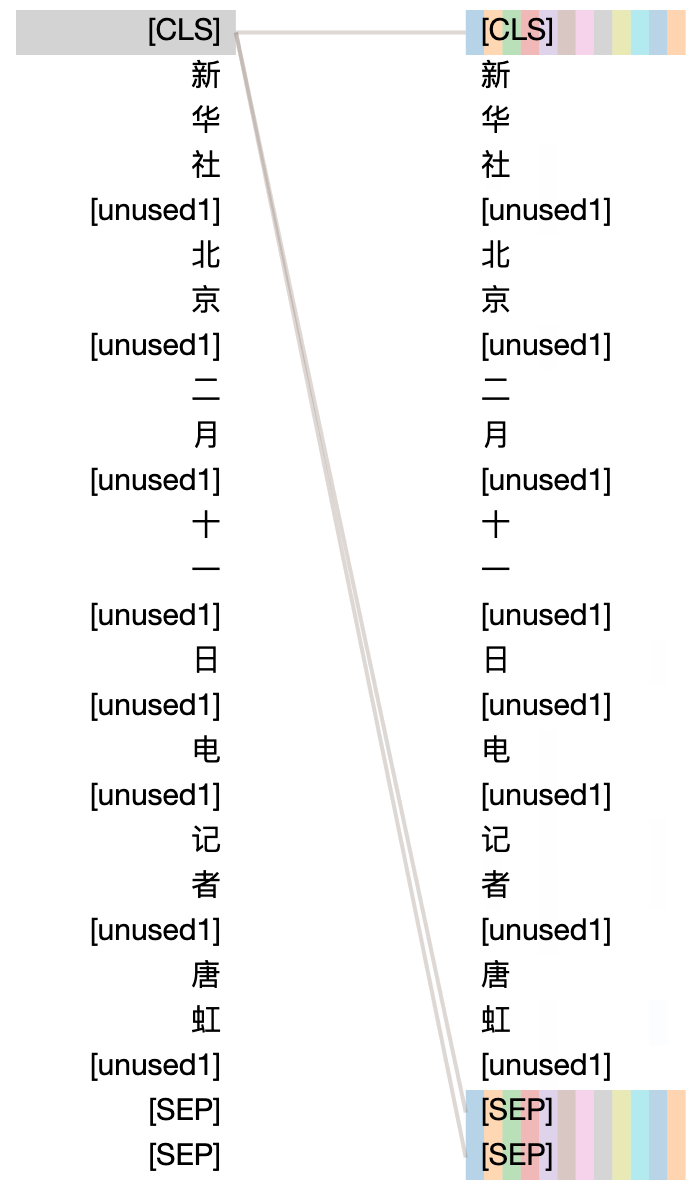}
\end{minipage}%
}%
\caption{Visualization of attentions of the markers selected from a random layer. We use $\texttt{[unused1]}$ in the BERT vocabulary as the inserted marker.}
\label{fig:vis}
\end{figure}

\subsubsection{MarkBERT -w/o marker}

Inserting markers is the key idea of solving the character and word dilemma in Chinese encoding.
In the NER task, inserting markers is important, indicating that MarkBERT structure is effective in learning word boundaries for tasks that requires such fine-grained representations.
In the NLU tasks, without inserting markers, MarkBERT-w/o marker can still achieve similar performances with the baseline methods, indicating that MarkBERT can also be used as a vanilla BERT model for easy usage in language understanding tasks.

\subsubsection{Visualization of Marker Attentions}

To further explore how the markers work in the encoding process, we use the attention visualization tool to show the attention weights of the inserted markers.
We explore the attention weights on the pre-trained MarkBERT and the fine-tuned model based on the Ontonotes NER task.
As seen in Figure \ref{fig:vis}, 
in some heads of the representations of the inserted markers, the attentions focus on the local semantics (e.g. in Fig. \ref{fig:vis} (a), the marker is attended to '二' (second) and '月'(month) in the head colored with purple and orange, indicating that the marker learn the context of the word '二月' (Feburary).
Further, the special tokens are the mostly focused as seen in Fig. \ref{fig:vis} (d).

\begin{figure}[]
\centering
\includegraphics[width=1.0\linewidth]{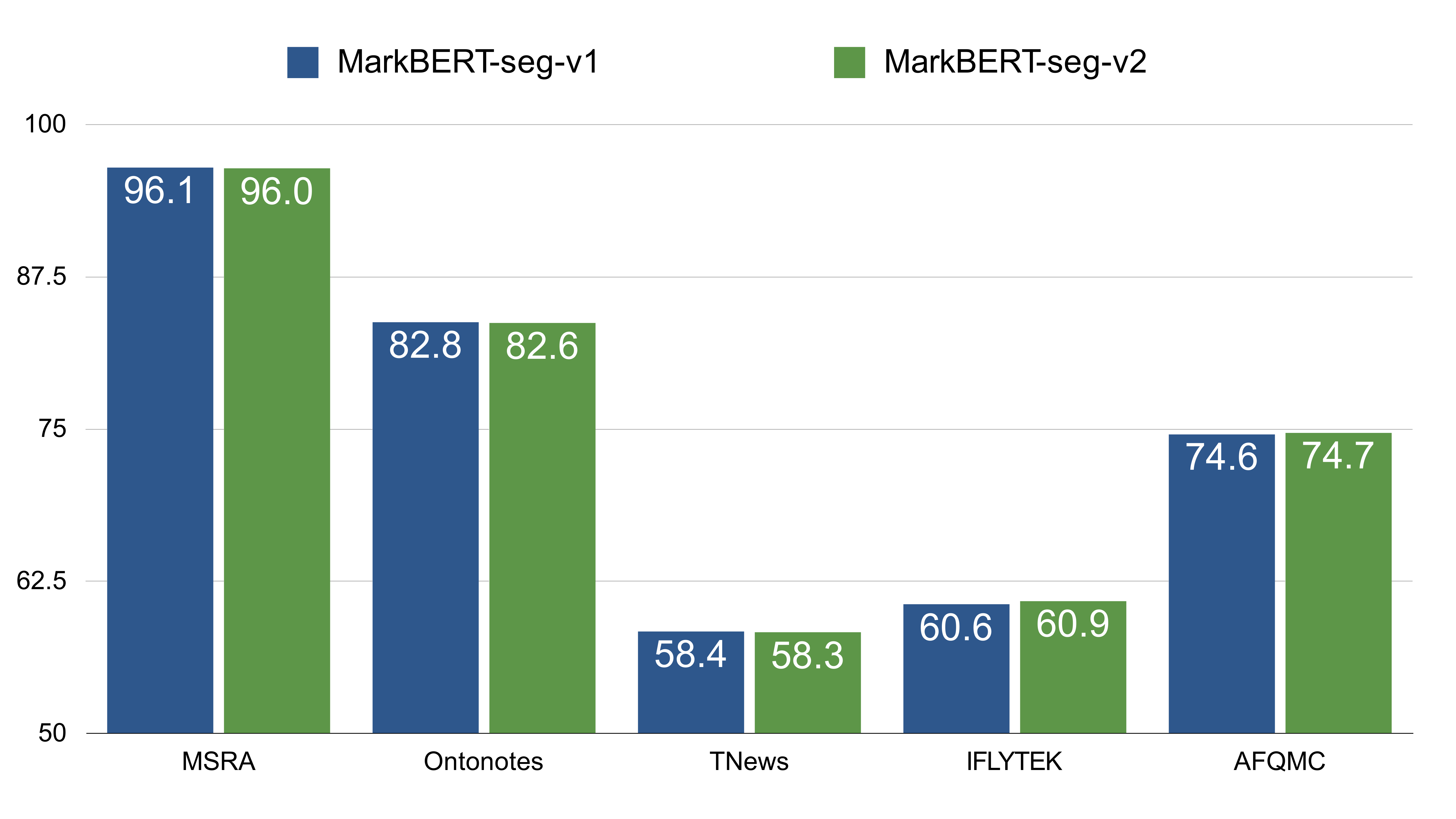}
\centering
\caption{Results on different MarkBERT versions.}
\label{fig:markbertcomparison}
\end{figure}

\subsubsection{Influence of Different Sementation Tools in MarkBERT}

The quality of the pre-processed segmentation results may play a vital role, therefore, we use a different version of segmentation in the Texsmart toolkit \cite{texsmart2020} where the segmentations are more fine-grained to train a MarkBERT-seg-v2 model as a comparison.

As seen in figure \ref{fig:markbertcomparison}, 
segmentation quality is trivial to MarkBERT.
The performances of MarkBERT (seg-v1) is similar to a variant MarkBERT-seg-v2 using a different segmentation tool,
which indicates that the training framework helps rather than the information from an external segmentation tool.

Combined with results in Table \ref{tab:bert-style}, we can conclude that introducing segmentation tools and use mark-style encoding is important while the quality of the segmentation is trivial. 

\section{Conclusion and Future Work}

In this paper, we have introduced MarkBERT, a simple framework for Chinese language model pre-training.
We insert special markers between word spans in the character-level encodings in pre-training and fine-tuning to make use of word-level information in Chinese.
We test our proposed model on the NER tasks as well as natural language understanding tasks.
Experiments show that MarkBERT makes significant improvements over baseline models.
In the future, we are hoping to incorporate more information to the markers based on the simple structure of MarkBERT.
\end{CJK*}

\bibliography{aaai22}
\end{document}